\documentclass[runningheads]{llncs}
\usepackage{graphicx}
\usepackage{mathtools,amstext,amsmath,amssymb}
\usepackage{multirow}
\usepackage{graphicx}
\usepackage{adjustbox}

\usepackage[colorlinks=true]{hyperref}
\usepackage{lmodern}
\usepackage{siunitx}
\usepackage{booktabs}
\usepackage{etoolbox}
\usepackage{soul}

\usepackage[font=small,skip=0pt]{caption}

\begin{document}
\title{Improved Inference via Deep Input Transfer
}
%
%

\author{Saied Asgari Taghanaki \and
Kumar Abhishek\and
Ghassan Hamarneh}


%
\authorrunning{Taghanaki et al.}
%
\institute{School of Computing Science, Simon Fraser University, Canada\\
\email{\{sasgarit, kabhishe, hamarneh\}@sfu.ca}}
\maketitle              
\begin{abstract}
Although numerous improvements have been made in the field of image segmentation using convolutional neural networks, the majority of these improvements rely on training with larger datasets, model architecture modifications, novel loss functions, and better optimizers. In this paper, we propose a new segmentation performance boosting paradigm that relies on optimally modifying the network's input instead of the network itself. In particular, we leverage the gradients of a trained segmentation network with respect to the input to transfer it to a space where the segmentation accuracy improves. We test the proposed method on three publicly available medical image segmentation datasets: the ISIC 2017 Skin Lesion Segmentation dataset, the Shenzhen Chest X-Ray dataset, and the CVC-ColonDB dataset, for which our method achieves improvements of 5.8\%, 0.5\%, and 4.8\% in the average Dice scores, respectively.

\end{abstract}
\section{Introduction}
Recently, there have been considerable advancements in semantic image segmentation using convolutional neural networks (CNNs), which have been applied to interpretation tasks on both natural and medical images~\cite{litjens2017survey}. The improvements are mostly attributed to exploring new neural architectures (with varying depths, widths, and connectivity or topology), designing new types of components or layers, adopting new loss functions, and training on larger datasets (via augmentation or acquisition). As one of the first high impact CNN-based segmentation models, Long et al.,~\cite{long2015fully} proposed fully convolutional networks for pixel-wise labeling. Next, encoder-decoder (and similarly convolution-deconvolution) segmentation networks were introduced~\cite{badrinarayanan2015segnet}. Soon after, Ronneberger et al.~\cite{ronneberger2015u} showed that adding skip connections to the segmentation network improves model accuracy and addresses vanishing gradients. More recent advancements include densely connected CNN architectures~\cite{jegou2017one}, learnable skip connections~\cite{taghanaki2018select}, hybrid object detection-segmentation~\cite{he2017mask}, and a new encoder-decoder architecture with Atrous separable convolution~\cite{chen2018encoder}.

Designing new loss functions also resulted in improvements in subsequent inference-time segmentation accuracy, e.g., optimizing various segmentation prediction metrics, such as the intersection over union~\cite{matthew2018lovasz} and the Dice score~\cite{milletari2016v}, controlling the level of false positives and negatives~\cite{taghanaki2018combo,wong20183d}, and adding regularizers to loss functions to encode geometrical and topological shape priors~\cite{aicha_shapepriors,mirikharaji2018star}.

Some previous works resorted to modifying the input image to improve the segmentation results. These modifications included applying conventional image normalization techniques prior to feeding the image to a segmentation network, e.g., Haematoxylin and Eosin pre-processing~\cite{cui2018deep}, edge-preserving smoothing~\cite{pal2015brief}, and whitening transformation~\cite{kannan2018segmentation}. Other works generated variants of the input image to augment the training data by applying radiometric and spatial image transformations, e.g., rotation, color shifting/normalization, and elastic deformation~\cite{shen2016automatic}. The shortcoming of such pre-processing methods is that they are not explicitly optimized to improve a specific task e.g., segmentation or classification. To the best of our knowledge, no previous work has optimized the manipulation of the input image in order to improve segmentation accuracy of a trained network. 
Recently, Drozdzal et al.~\cite{drozdzal2018learning} showed that attaching a pre-processing module at the beginning of a segmentation network improves the network performance. However, we argue that adding a pre-processor without any other explicit constraint(s) amounts to adding (or prefixing) more layers, i.e., essentially making the model deeper. Inspired by adversarial perturbations~\cite{kurakin2016adversarial,xie2017adversarial}, in this paper, we choose to optimally modify the input image prior to feed-forward inference. Our input-transformation is carried out via a novel gradient based method that leverages the computational processes of any trained segmentation network. After calculating these optimal transformations on training data, we then learn an image-to-image translation network that estimates an image modification mapping for novel \textit{test} images. Note that our input transformation is \emph{not} a data-augmentation method (albeit data augmentation may still be performed independently of our method), rather, we learn (from training data) a translation network that will pre-process novel input at inference time.

In this paper, we make the following contributions: a) we introduce the first iterative gradient-based input pre-processing method, b) we adopt an explicit objective to effectuate the purpose of the pre-processor, and c) we show how targeted gradient-based adversarial perturbation methods can be leveraged for a better segmentation performance.

\section{Method}\label{sec:method}

Segmenting an input image $\mathbf{I}$ of size $n \times m$ assigns a label $l_i \in \mathcal{C} = \{0,1,\cdots,L-1\}$ to the each pixel in $\mathbf{I}$, where $L$ is the number of classes. Given a segmentation network with parameters  $\mathbf{\Theta}$, let $f(\mathbf{I; \Theta})$ denote the pixel-wise activation for $\mathbf{I}$ before the softmax normalization (denoted by $\xi_{\mathcal{C}}$), and let $\mathbf{\hat{\mathcal{S}}} \in \mathbb{R}^{n \times m \times L}$ represent the segmented image as,
\begin{equation}
    \xi_{\mathcal{C}} \left( f (1-\mathbf{\mathbf{\mathcal{S}}(I + \mathbf{\Delta}_{\mathbf{I}}); \Theta}) \right) = \mathbf{\hat{\mathcal{S}}}.
\end{equation}
Let $\mathbf{\mathcal{S}} \in \mathbb{R}^{n \times m \times L}$ denote the ground truth segmentation for $\mathbf{I}$. For a perfect segmentation, $\mathbf{\hat{\mathcal{S}}} = \mathbf{\mathcal{S}}$. Our goal is to introduce a perturbation $\mathbf{\Delta}_{\mathbf{I}}$ to $\mathbf{I}$, such that the segmentation output of the modified image $\mathbf{I + \mathbf{\Delta}_{\mathbf{I}}}$ is equal to the ground truth, i.e.,
\begin{equation}
    \xi_{\mathcal{C}} \left( f (\mathbf{\mathcal{S}}(\mathbf{I + \mathbf{\Delta}_{\mathbf{I}}); \Theta}) \right) = \mathbf{{\mathcal{S}}}
\end{equation}
Let $f_{\mathbf{\hat{\mathcal{S}}}}(\mathbf{I; \Theta})$ and $f_{\mathbf{{\mathcal{S}}}}(\mathbf{I; \Theta})$ represent the pixel-wise activations corresponding to the segmentation outputs $\hat{\mathcal{S}}$ and ${\mathcal{S}}$ respectively. We apply a gradient descent algorithm for estimating the perturbation $\mathbf{\Delta}_{\mathbf{I}}$ to be added to $\mathbf{I}$. The objective function $\mathcal{G}$ for this can be written as 
\begin{equation}
    \mathcal{G}(\mathbf{I, \hat{\mathcal{S}}, \mathcal{S}, \Theta}) = f_{\mathbf{\hat{\mathcal{S}}}}(\mathbf{I + \mathbf{\Delta}_{\mathbf{I}}, \Theta}) - f_{\mathcal{S}}(\mathbf{I +\mathbf{\Delta}_{\mathbf{I}}, \Theta}).
\end{equation}
Starting with the original image $\mathbf{I}$, we iteratively compute the gradient of the loss $\mathcal{G}(\mathbf{I, \hat{\mathcal{S}}, \mathcal{S}, \Theta})$ with respect to $\mathbf{I} + \pmb{\delta}'^{(k)}_{\mathbf{I}}$ and add it to the image. Note that $\pmb{\delta}'^{(k)}_{\mathbf{I}}$ is zero for the first iteration.  Let $\mathbf{I}^{(k)}$ denote the perturbed image after the $k^{th}$ iteration of gradient descent. For the $k^{th}$ iteration, we have 
\begin{equation} \label{eq:normalization}
    \mathbf{I}^{(k+1)} = \mathbf{I}^{(k)} + \gamma \ \pmb{\delta}'^{(k)}_{\mathbf{I}}
\end{equation}
where $\gamma$ is a scaling constant, and $\pmb{\delta}'^{(k)}_{\mathbf{I}}$ is the gradient obtained for the $k^{th}$ iteration calculated by gradient descent update as \\
\begin{equation} \label{eq:pert_calc}
\begin{split}
    \pmb{\delta}^{(k)}_{\mathbf{I}} &= 
    \nabla \mathcal{G}(\mathbf{I, \hat{\mathcal{S}}, \mathcal{S}, \Theta})\\&=  \nabla_{\mathbf{I}^{(k)}} f_{\hat{\mathcal{S}}}(\mathbf{\mathbf{I + \mathbf{\Delta}_{\mathbf{I}}}, \Theta}) - \nabla_{\mathbf{I}^{(k)}} f_{\mathbf{\mathcal{S}}}(\mathbf{I + \mathbf{\Delta}_{\mathbf{I}}, \Theta})
\end{split}
\end{equation} 
and then normalized using its $L_\infty$ norm for numerical stability as 
\begin{equation}
    \pmb{\delta}'^{(k)}_{\mathbf{I}} = \frac{\pmb{\delta}^{(k)}_{\mathbf{I}}}{\|\pmb{\delta}^{(k)}_{\mathbf{I}}\|_\infty}.
\end{equation} 
The algorithm terminates if the segmentation of the modified image is equal (or close enough within an error margin) to the ground truth, or it reaches a certain maximum number of iterations $K$. The total perturbation for image $\mathbf{I}$ is then calculated as the sum of the individual perturbations, i.e., $\mathbf{\Delta}_{\mathbf{I}} = \sum_k \pmb{\delta}'^{(k)}_{\mathbf{I}}$. The segmentation output of this perturbed image $\mathbf{X + \Delta_{\mathbf{I}}}$ is denoted by $\mathcal{S}^*$.

\begin{figure*}[]
\centering
  \includegraphics[width=1\textwidth]{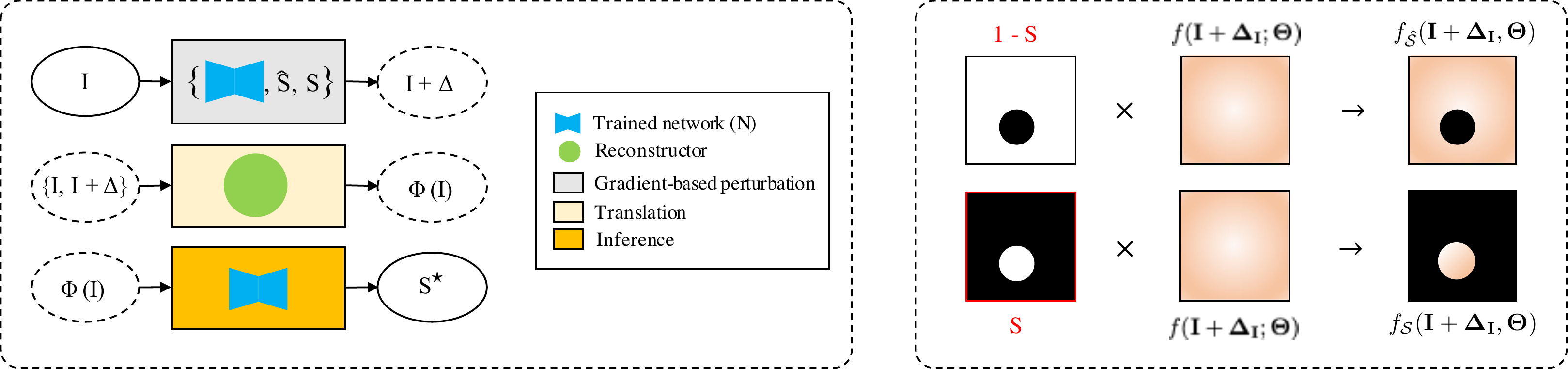}
  \caption{Left: Passing image $\mathbf{I}$ through the trained network $N$ generates sub-optimal output $\hat{\mathcal{S}}$. Gradient perturbation module (top) calculates perturbation $\mathbf{\Delta}$ on $\mathbf{I}$, such that passing $\mathbf{I}+\mathbf{\Delta}$ through $N$ yields results closer to ground truth $S$. Translation network (middle) is trained to learn mapping $\mathbf{\Phi:I \rightarrow  I+\Delta}$. Test images are first transformed via $\mathbf{\Phi}$ before feeding them into $N$ (bottom), which results in an improved segmentation output $\mathcal{S}^*$. Right: Reducing the logits of background and increasing that of foreground.}
  \label{method-diagram}
\end{figure*}

The calculation of $\mathbf{\Delta}_{\mathbf{I}}$ requires knowledge of the ground truth segmentation mask and thus is only available for the training data. To test the hypothesis that the proposed gradient-based method improves the segmentation performance, we perturb the test images with their corresponding ground truths, and we obtain an almost perfect segmentation (Dice score $\sim 1.0$). However, since the ground truth segmentation masks for test images are not available in practice, we propose to reconstruct an estimate of the perturbed test images. In particular, 
given pairs of training images and their corresponding perturbations, $\left\{\left(\mathbf{I}_{train}, \mathbf{I}_{train} + \ \mathbf{\Delta}_{\mathbf{I}_{train}}\right)\right\},$ we train a deep model to learn  $\mathbf{\Phi}: \mathbf{I}_{train} \to \mathbf{I}_{train} + \ \mathbf{\Delta}_{\mathbf{I}_{train}}$. Subsequently, we apply the learned $\mathbf{\Phi}$ to the test data to obtain the reconstruction $\mathbf{I}_{test} \to \mathbf{I}_{test} + \ \mathbf{\Delta}_{\mathbf{I}_{test}}$.
To learn 
the transformation function $\mathbf{\Phi}(\cdot)$, an image-to-image translation network can be used. Figure~\ref{method-diagram} shows an overview of the proposed method.

\section{Implementation Details and Data}   \label{sec:details}
\subsection{Models} \label{subsec:models}
As the goal of this work is to demonstrate the effectiveness of the proposed gradient-based perturbation, we use a state-of-the-art baseline segmentation network i.e., the U-Net~\cite{ronneberger2015u} and optimize it using Adadelta with a batch size of 64. To learn the transformations made by the proposed gradient-based perturbation method (GP) from the training data as discussed in Section~\ref{sec:method}, we use two image-to-image translation networks: Cycle-GAN (cG)~\cite{zhu2017unpaired} and the hundred-layers Tiramisu segmentation network (T)~\cite{jegou2017one}. We modify the latter as an image-to-image translation network and replace the original Tiramisu network's loss function with a loss function consisting of two terms: a Structural Similarity Index Measure (SSIM) term and a mean absolute error (L1 loss) term:
\begin{gather}\label{eq:newLoss}
    \mathcal{L} = \sum_{\mathbf{I}_{train}} \Big[\left( 1 - \mathcal{SSIM}\left(\mathbf{I},\ \mathbf{I} + \mathbf{\Delta}\right) \right) + \lambda \ \|\mathbf{I},\ \mathbf{I} + \mathbf{\Delta}\|_1 \Big]
\end{gather}
where $\mathcal{SSIM}\left(\mathbf{I},\ \mathbf{I} + \mathbf{\Delta}\right)$ is the SSIM calculated between $\mathbf{I}$ and $\mathbf{I} + \mathbf{\Delta}$, and $\lambda$ is a scaling constant. The SSIM loss captures the finer perceptual details to which the human visual system is sensitive, such as contrast, luminance, and structure for which the L1 loss fails. When reporting results, we use the following abbreviations (i) ORIG: the original U-Net; (ii)  $\mathrm{GP_{cG}}$: the proposed GP ($\gamma = 1.0$) + Cycle-GAN for reconstructing the test image perturbation; (iii) $\mathrm{GP_T}$: GP ($\gamma=0.1$) + Tiramisu reconstruction with L1 loss; and (iv) $\mathrm{GP_{Ts}}$: GP ($\gamma=1.0$) + Tiramisu reconstruction with SSIM loss (s; Eqn.~\ref{eq:newLoss}). We choose the maximum possible value of $\gamma=1.0$ in Eqn.~\ref{eq:normalization} for $\mathrm{GP_{Ts}}$ and $\mathrm{GP_{cG}}$ methods to maximally perturb images with the goal of highest possible segmentation performance, and run the optimization for $K=100$ iterations. Because the perturbations can have negative values, we use a linear activation function for the last layer in all the aforementioned image-to-image translation networks.

\subsection{Data}
We use three datasets used to evaluate our method a) \textbf{The ISIC 2017 Skin Lesion Segmentation Dataset}~\cite{codella2017skin}, hereafter referred to as SKIN, consists of 2000 skin lesion images for training and 600 test images. For all the images, the lesions have been manually annotated by expert dermatologists for normal skin and other miscellaneous structures. b) \textbf{The Shenzhen Chest X-Ray Dataset}~\cite{Shenzhen}, hereafter referred to as LUNG, consists of 662 frontal chest X-Ray images, out of which 336 are cases with manifestations of tuberculosis and the remaining 326 are non-diseased ones. The corresponding ground truth masks contain manually traced out boundaries for the left and the right lungs. c) \textbf{The CVC-ColonDB}~\cite{bernal2012towards}, hereafter referred to as COLON, is a database of frames along with the corresponding annotated polyp masks extracted from colonoscopy videos. The dataset consists of 300 training images and 50 test images.

\section{Results and Discussion}    \label{sec:results}
\begin{figure}[]
  \centering
  \includegraphics[width=.9\textwidth]{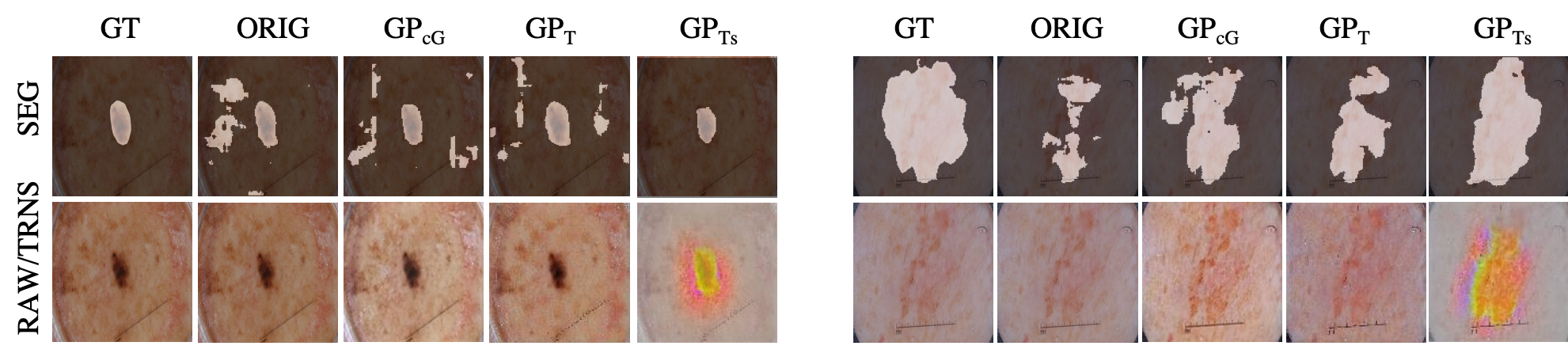}
  \caption{Segmentation results on 2 sample images from SKIN. The top row shows the segmentation masks overlaid on top of the image, while the bottom row shows the original images (RAW) in the first two columns and the transformed images (TRNS) in the next three columns.}
  \label{skin}
\end{figure}

\begin{figure}[h!]
  \centering
  \includegraphics[width=.9\textwidth]{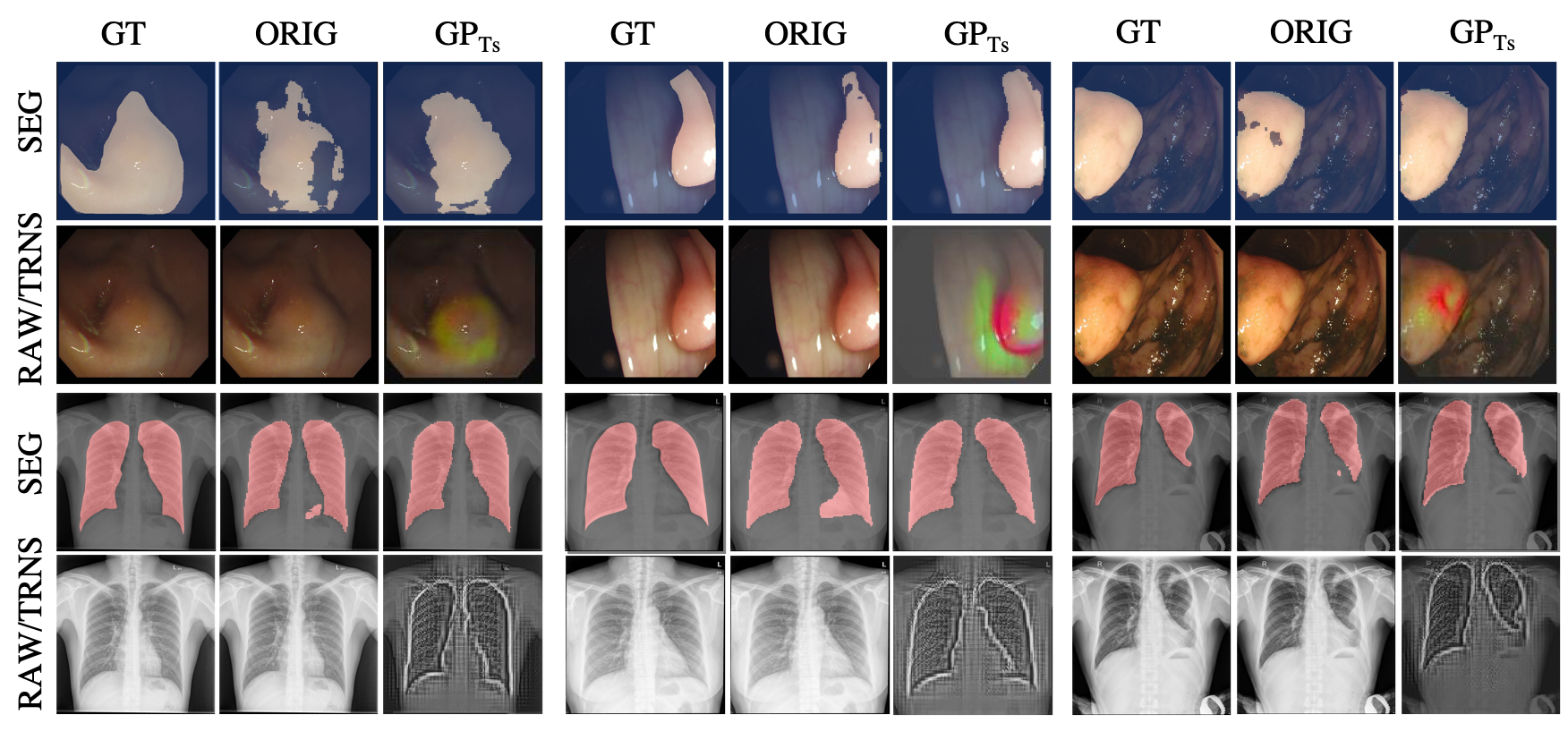}
  \caption{Sample segmentation and transformed results for the COLON (rows 1 and 2) and LUNG (rows 3 and 4) datasets. RAW and TRNS refer to normal and transformed images, respectively.}
  \label{colon_lung}
\end{figure}

\begin{figure}[h!]
\centering
\begin{tabular}{ccc}
\includegraphics[width=.265\textwidth]{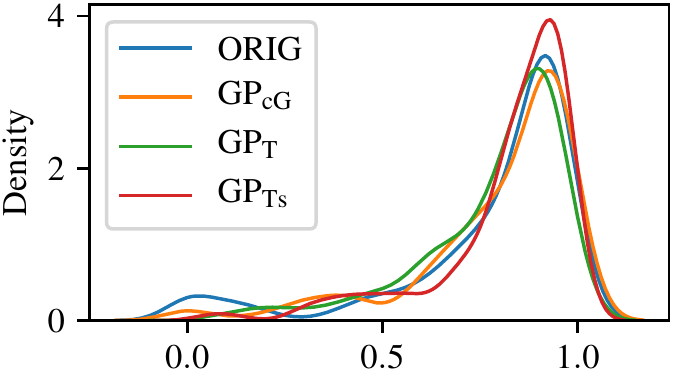} &\includegraphics[width=.25\textwidth]{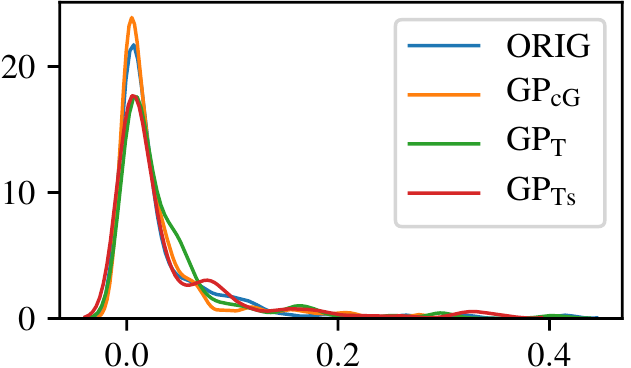} & \includegraphics[width=.25\textwidth]{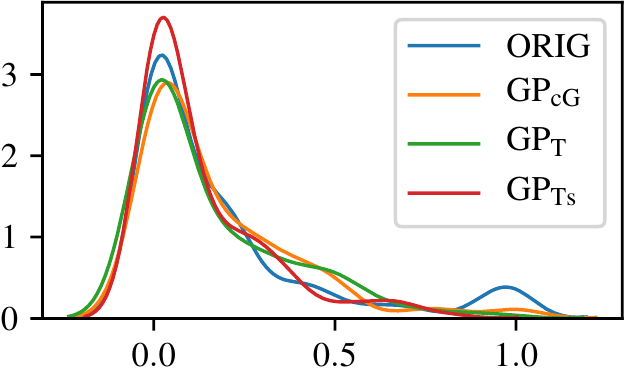} \\
\includegraphics[width=.273\textwidth]{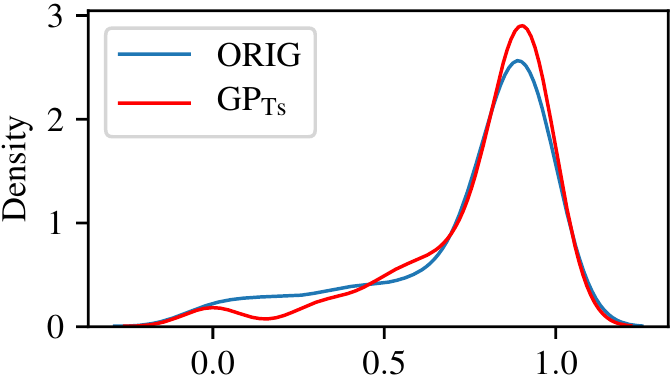} & \includegraphics[width=.25\textwidth]{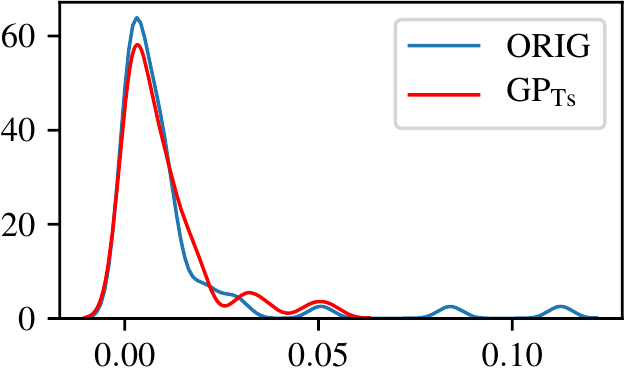} & \includegraphics[width=.25\textwidth]{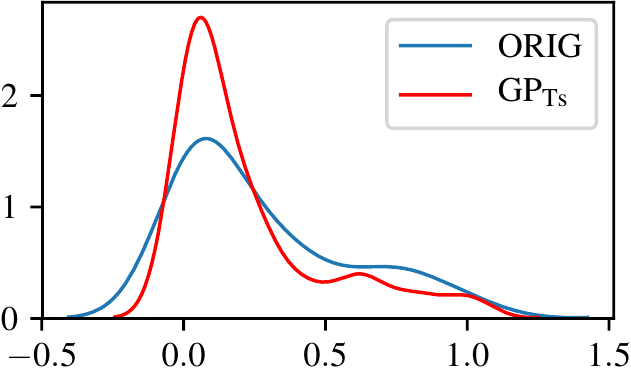} \\
\includegraphics[width=.273\textwidth]{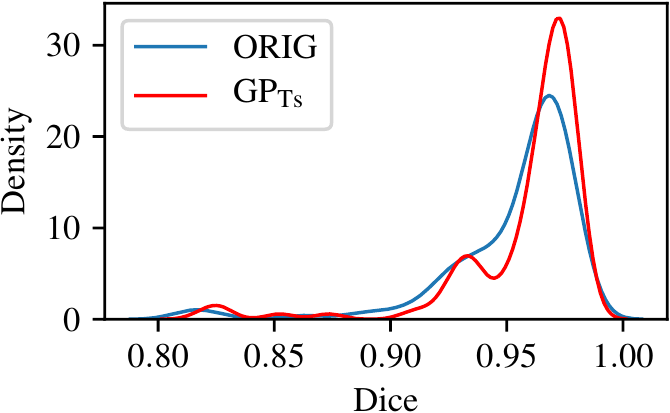} &  \includegraphics[width=.25\textwidth]{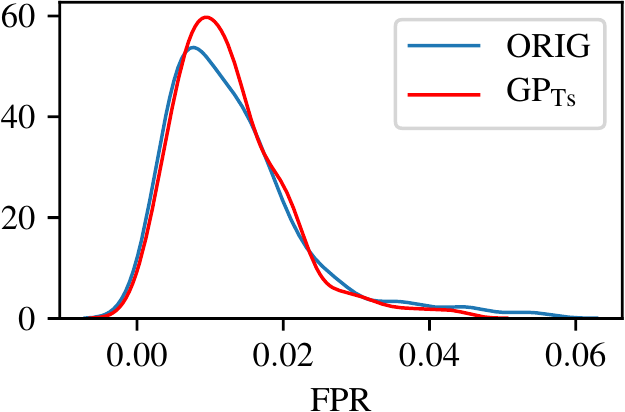} & \includegraphics[width=.25\textwidth]{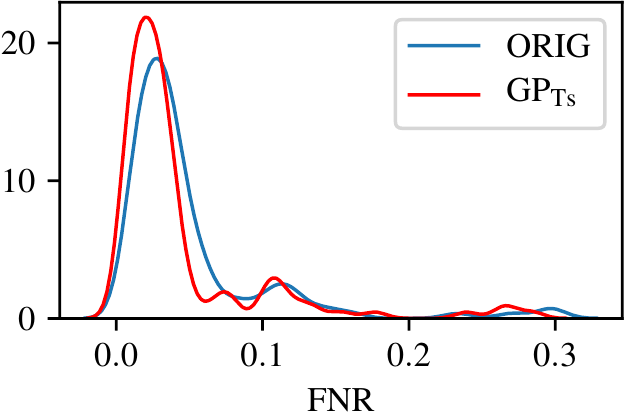} 
\end{tabular}
\caption{Kernel density estimation; 1st row: SKIN; 2nd row: COLON; 3rd row: LUNG.}
\label{KDE}
\end{figure}

To test the effectiveness of the proposed transformation method, in this section, we present both quantitative and qualitative results on the three aforementioned datasets. We start with SKIN and the different derivatives of the proposed gradient-based perturbation method as discussed in Section~\ref{subsec:models}, i.e., $\mathrm{GP_{cG}}$, $\mathrm{GP_T}$, and $\mathrm{GP_{Ts}}$ and compare them to ORIG. As shown in Figure~\ref{skin}, the proposed method (i.e., $\mathrm{GP_{Ts}}$) produces the closest segmentation mask to the ground truth (GT) compared to other methods. Looking at the second row of Figure~\ref{skin}, $\mathrm{GP_{Ts}}$ perturbs the pixel values in a band surrounding the lesion, enhancing its contrast, making it more distinguishable from the background, thereby boosting the segmentation network's result, especially around the critical lesion boundary pixels. 
\vspace{-15pt}
\smallskip
\vspace{-20pt}
\noindent In Table~\ref{colon_lung}, we report the quantitative results for SKIN obtained with U-Net. As shown, all the gradient-based  perturbation methods outperform ORIG, with the proposed $\mathrm{GP_{Ts}}$ method achieving a significant improvement of the mean Dice score by 5.8\% compared to ORIG.
Moreover, a visual inspection of the ORIG and the $\mathrm{GP_{Ts}}$ results (Figure~\ref{skin}, right) shows that the latter is more adept at rejecting false negative pixels, which is also supported by the results from Table~\ref{colon_lung} where $\mathrm{GP_{Ts}}$ improves the False Negative Rate (FNR) by a considerably large amount (5.87\%).
Next, we pick the best-performing method from SKIN experiments, i.e., $\mathrm{GP_{Ts}}$, and evaluate its performance on the remaining two datasets: COLON and LUNG. Figure~\ref{colon_lung} shows the qualitative results obtained for ORIG and $\mathrm{GP_{Ts}}$ applied to three sample images from LUNG and COLON. As can be seen from the results, $\mathrm{GP_{Ts}}$ obtains segmentation results much closer (i.e., smoother with no perforated or disconnected blobs) to the ground truth segmentation (GT).
\vspace{-15pt}
\vspace{-20pt}
This is also validated by the quantitative results reported in Table~\ref{lung_colon_quant} where $\mathrm{GP_{Ts}}$ outperforms ORIG in all the three metrics - Dice score, FPR, and FNR, obtaining 4.8\% and 0.5\% improvements in the mean Dice score for COLON and LUNG, respectively.
\vspace{-20pt}
\begin{table}[]
\sisetup{detect-weight,mode=text}
\renewrobustcmd{\bfseries}{\fontseries{b}\selectfont}
\renewrobustcmd{\boldmath}{}
\newrobustcmd{\B}{\bfseries}
\addtolength{\tabcolsep}{-4.1pt}
\setlength{\tabcolsep}{6pt}
\centering
\caption{SKIN, LUNG and COLON segmentation results; Mean $\pm$ standard error.}
\begin{tabular}{llccc}
\hline
Data & Method & Dice & FPR & FNR \\ \hline
\multirow{2}{*}{SKIN} & ORIG & 0.7743 $\pm$ 0.020 & 0.0327 $\pm$ 0.005 & $0.1905 \pm 0.022$ \\
 & $\mathrm{GP_{cG}}$ & 0.7928 $\pm$ 0.017 & \textbf{0.0281} $\pm$ \textbf{0.002} & 0.1759 $\pm$ 0.014 \\
 & $\mathrm{GP_{T}}$ & 0.7836 $\pm$ 0.015 & 0.0396 $\pm$ 0.005 & 0.1595 $\pm$ 0.017 \\
 & $\mathrm{GP_{Ts}}$ & \textbf{0.8190} $\pm$ \textbf{0.015} & 0.0399 $\pm$ 0.006 & \textbf{0.1318} $\pm$ \textbf{0.014}\\ \hline
\multirow{2}{*}{LUNG} & ORIG & 0.9527 $\pm$ 0.003 & 0.0133 $\pm$ 0.0010 & 0.0504 $\pm$ 0.005 \\
 & $\mathrm{GP_{Ts}}$ & \textbf{0.9578} $\pm$ \textbf{0.003} & \textbf{0.0126} $\pm$ \textbf{0.0007} & \textbf{0.0441} $\pm$ \textbf{0.005} \\ \hline
\multirow{2}{*}{COLON} & ORIG & 0.7384 $\pm$ 0.040 & 0.0115 $\pm$ 0.0030 & 0.2799 $\pm$ 0.042 \\
 & $\mathrm{GP_{Ts}}$ & \textbf{0.7737} $\pm$ \textbf{0.033} & \textbf{0.0099} $\pm$ \textbf{0.0020} & \textbf{0.2235} $\pm$ \textbf{0.038} \\ \hline
\end{tabular}
\label{lung_colon_quant}
\end{table}
\vspace{-15pt}
To further capture the improvement in segmentation performance, we plot the Gaussian kernel density estimates to estimate the probability density functions of Dice score, FPR, and FNR for the three datasets. The plots in Figure~\ref{KDE} support the quantitative results with higher peaks (which correspond to higher densities) at larger Dice values for $\mathrm{GP_{Ts}}$ as compared to ORIG for all three datasets. Next, we look at range of the three metrics, and observe that they are in general more restricted to higher values for Dice score and lower values for FPR and FNR for $\mathrm{GP_{Ts}}$ than ORIG.
Although we achieve up to $\sim 5\%$ improvement in Dice score, it is important to note that a much larger possible improvement is lost during the reconstruction phase. For example, for SKIN, when we perturb the test images with their corresponding ground truths as described in Section~\ref{sec:method}, we obtain almost perfect segmentation results i.e. Dice $\sim 1.0$ (we emphasize that the performance improvement is solely from the perturbation and not from the image-to-image translation components), but the best results we obtain through reconstruction is far less. We validated this by calculating the SSIM between the $\mathrm{GP}$ and the $\mathrm{GP_{Ts}}$ images, and it was $0.32 \pm 0.009$. This shows that a lot of information is lost in the reconstruction phase. Since training our input-perturbing mechanism requires gradients from an already-trained segmentation network, along with pixel-level class labels,  training our perturbation module and training the segmentation network cannot be done simultaneously. In our next experiment, we set out to demonstrate that using (i) our pre-trained perturbation network as a pre-processor to a segmentation network leads to better results than (ii) an end-to-end training of the pre-processor network serially connected to the segmentation network. In other words, we compare the segmentation performance when training the pre-processor with (i) vs. without (ii) the proposed gradient based constraint. We find that the proposed method (i) achieves a higher Dice score of 0.8190 compared to 0.8019 using (ii).
\vspace{-15pt}
\vspace{-40pt}
\section{Conclusion}    \label{sec:conclusion}
We proposed a novel input image transformation optimized for improved segmentation. A network gradient-based method calculates signed input perturbations on training images, which are then used to train deep networks to infer test image transformations. Our evaluations showed that the proposed approach can improve the performance of baseline methods for a variety of medical imaging modalities. A direction for future work includes focusing on improving the translation step of the proposed method. Moreover, the proposed method can be extended to other tasks such as image classification and object detection.

%
%
%
\bibliographystyle{splncs04}
\bibliography{samplepaper.bbl}
%




\end{document}